\documentclass[a4paper,longmktitle]{cas-sc}
\usepackage{natbib}
\setcitestyle{square,comma,numbers, longnamesfirst}

\usepackage{amsmath}
\usepackage{mathrsfs}
\usepackage{graphics}
\usepackage{subfigure}
\usepackage{xcolor}

\def\tsc#1{\csdef{#1}{\textsc{\lowercase{#1}}\xspace}}
\tsc{WGM}
\tsc{QE}
\tsc{EP}
\tsc{PMS}
\tsc{BEC}
\tsc{DE}

\begin{document}
	\let\WriteBookmarks\relax
	\def\floatpagepagefraction{1}
	\def\textpagefraction{.001}
	\shorttitle{NPMCA-net}
	\shortauthors{Siyue Yu et~al.}
	
	\title [mode = title]{Fast Pixel-Matching for Video Object Segmentation}                      
	
	

	\author[1]{Siyue Yu}[style=chinese]
	\ead{siyue.yu@xjtlu.edu.cn}

	\address[1]{Xi'an Jiaotong-Liverpool University, Suzhou, Jiangsu, China}
	
	\author[1]{Jimin Xiao}[style=chinese]
	\cormark[1]
	\ead{jimin.xiao@xjtlu.edu.cn}
	
	\author[1]{Bingfeng Zhang}[style=chinese]
	\ead{bingfeng.zhang@xjtlu.edu.cn}
	
	
	
	\author%
	[1]{Eng Gee Lim}[style=english]
	\ead{enggee.lim@xjtlu.edu.cn}
	
		
	\author%
	[2]{Yao Zhao}[style=chinese]
	\ead{yzhao@bjtu.edu.cn}
	
	
	\address[2]{Beijing Jiaotong University, Beijing, China}
	
	\cortext[cor1]{Corresponding author}
	\nonumnote{The work was supported by National Natural Science Foundation of China under 61972323, and Key Program Special Fund in XJTLU under KSFT-02, KSF-P-02.}
	

\begin{abstract}
Video object segmentation, aiming to segment the foreground objects given the annotation of the first frame, has been attracting increasing attentions. Many state-of-the-art approaches have achieved great performance by relying on online model updating or mask-propagation techniques. However, most online models require high computational cost due to model fine-tuning during inference. Most mask-propagation based models are faster but with relatively low performance due to failure to adapt to object appearance variation. In this paper, we are aiming to design a new model to make a good balance between speed and performance. We propose a model, called NPMCA-net, which directly localizes foreground objects based on mask-propagation and non-local technique by matching pixels in reference and target frames. Since we bring in information of both first and previous frames, our network is robust to large object appearance variation, and can better adapt to occlusions. Extensive experiments show that our approach can achieve a new state-of-the-art performance with a fast speed at the same time (86.5$\%$ IoU on DAVIS-2016 and 72.2$\%$ IoU on DAVIS-2017, with speed of 0.11s per frame) under the same level comparison. Source code is available at \url{https://github.com/siyueyu/NPMCA-net}.  

\end{abstract}



	
	
	

\begin{keywords}
	non-local pixel matching \sep mask-propagation \sep encoder-decoder
\end{keywords}

\maketitle

\section{Introduction}


Video object segmentation (VOS) has been attracting increasing attention in recent years due to its significance in video understanding. The aim of this task is to track the target object from the first frame to the end of the video sequence and segment all the pixels belonging to the tracked target object, which faces problems of object occlusion and appearance variance.

To tackle these problems, some studies adopted online-training mechanism ~\cite{Caelles_2017_CVPR, maninis2018video, voigtlaender2017online, perazzi2017learning}. Given the ground-truth mask of the first frame in a test video, they used it to fine-tune the model to obtain the object appearance. In the following inference process, they used the predicted masks to further fine-tune their models. With fine-tuning, the models can adapt to object appearance change, though, the online learning process is time-consuming and inefficient.

Recently, boosted by the rapid development of mask-propagation based VOS models ~\cite{wug2018fast,johnander2019generative,lin2019agss}, a better balance between speed and accuracy is reached. The core idea of these methods is to use the estimated mask of the previous frame to guide the model to make segmentation prediction for the current frame. For example, Perazzi et al.~\cite{perazzi2017learning} proposed to use guidance of previous predicted mask as guidance for the network to learn mask prediction and it proposed a combination of offline and online training method to train the model. They firstly used static image datasets for offline training, and then used the first frame of a test video sequence to fine-tune the model. Oh et al.~\cite{wug2018fast} proposed a Siamese encoder-decoder network with guidance of the previous mask to produce the target object probability map. Johnander et al.~\cite{johnander2019generative} offered an appearance module which utilized a class-conditional mixture of Gaussians to model the foreground object appearance for mask prediction. Sun et al.~\cite{sun2018mask} considered both the mask of previous frame and the optical flow to predict target mask. These approaches are usually faster than online training based VOS methods, but they are less adaptive to object appearance variation.

Both online training and mask-propagation based VOS models have limitations, a balance between segmentation accuracy and running speed is crucial for VOS. Early mask-propagation based networks use current frame with previous estimated mask~\cite{perazzi2017learning} or adding first frame with its provided mask as reference information~\cite{wug2018fast} to directly predict the segmentation mask of current frame. Additionally, Sun et al.~\cite{sun2018mask} used optical flow to build relationship between the previous and the current frames. Different from these methods, we design an attention-based pixel-matching module to find the pixels belonging to the target object in the current frame based on the feature similarity between the current frame and reference frames. In order to capture the object feature without the interference of background, we choose to mask it out and discard the background pixels. However, the target object is varying frame by frame, such process will cause large object appearance variation. Therefore, we choose to use both the first frame and the previous frame as references to provide object information for our pixel-matching module.

With the target object's appearance information, we need to determine the target object location, in terms of mask, in the current frame. We design our model based on mask-propagation to keep efficiency, where the non-local structure~\cite{Wang_2018_CVPR} is adopt to generate the object mask using the obtained target object's appearance information. Specifically, we design a video object segmentation model called Non-local Pixel-Matching network with Channel Attention (NPMCA-net), which includes a newly designed pixel-matching module and a channel attention module. The pixel-matching module is designed to match pixels between the target frame and the reference frames with given ground-truth mask or estimated mask. The channel attention module is used to augment the matched feature map to achieve better decoding. Extensive experiments have shown that our network can achieve a new state-of-the-art performance without loss of efficiency. To better display the accuracy and speed trade-off, we plot our IoU score versus speed in Fig.~\ref{FIG:speed}. Our NPMCA-net can achieve both high performance and high efficiency at the same time. Our main contribution is summarized as follows:

\begin{figure*}
	\centering
		\includegraphics[scale=.5]{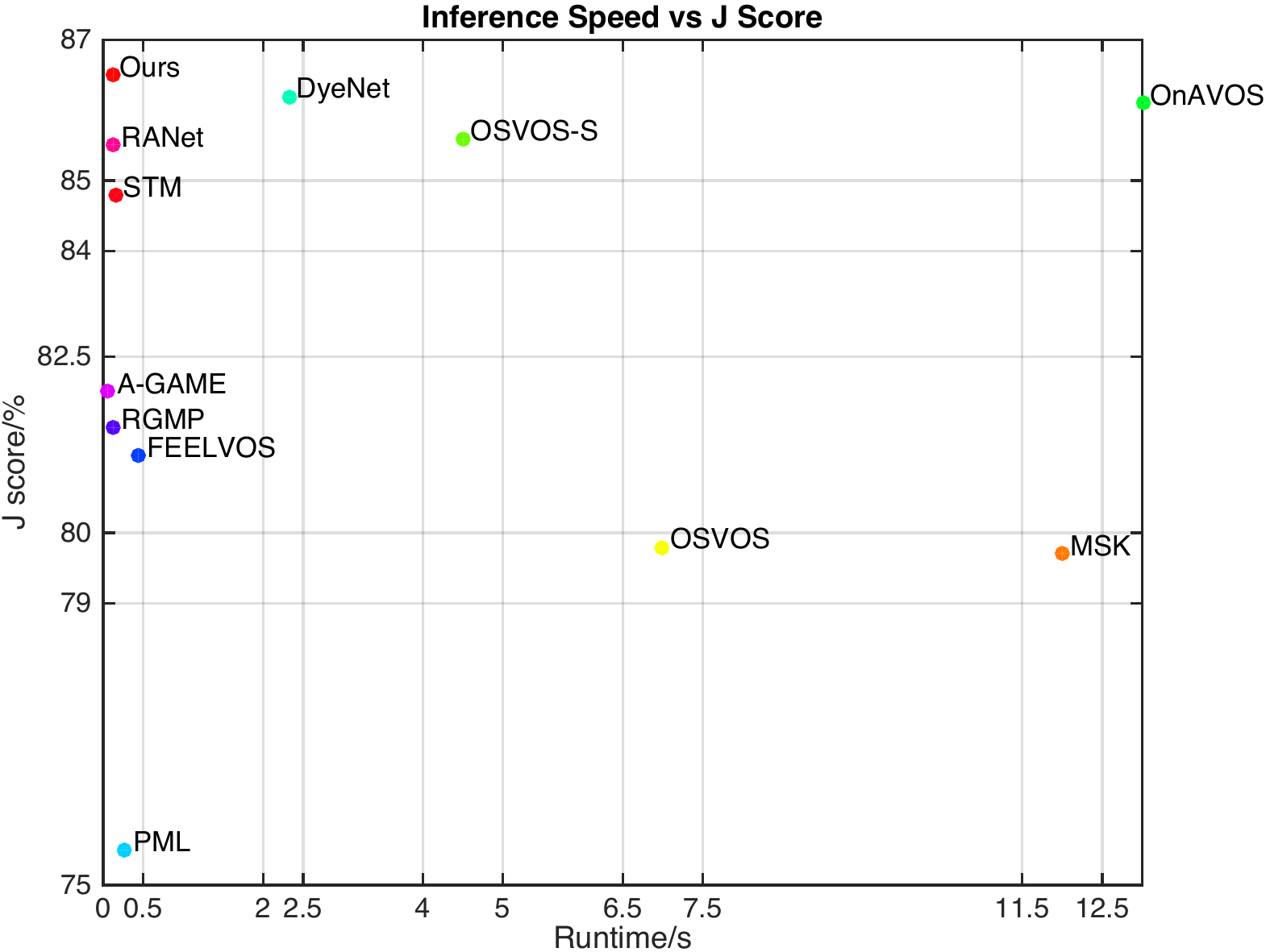}
	\caption{The IoU score ($\mathcal{J}$) versus running time on each frame ($s$) for various VOS approaches on the DAVIS-2016 validation set. Our model can keep a good balance between performance and efficiency.}
	\label{FIG:speed}
\end{figure*}

\begin{itemize}
    \item We propose a video object segmentation model (NPMCA-net) that strikes a good balance between accuracy and running speed. The model does not rely on online fine-tuning technique, so as to lower the computational demands, yet it can adaptively catch the target object's appearance variation by using both image and predicted mask information in the previous frame.  
    
    \item Our proposed non-local pixel-matching module can effectively predict the target object mask by aggregating multi-frame information. Moreover, the proposed model also provides high level interpretability by visualizing the obtained feature maps. 
    
    \item Our model achieves new state-of-the-art performances on DAVIS-2016 (IoU: 86.5$\%$) and DAVIS-2017 (IoU: 72.2$\%$) datasets, using the same experimental setting. 
    
\end{itemize}

\section{Related Works}
\textbf{Video Object Segmentation.}
Different from statistic image tasks ~\cite{chen2017deeplab, Liu_2019_CVPR, zhang2019reliability, zhang2021self, yu2020structure}, VOS only considers to segment the moving object without class reference or prediction. VOS research can be divided into two main categories, one is unsupervised methods and the other is semi-supervised methods. Unsupervised methods, such as~\cite{Wang_2019_CVPR,Li_2018_CVPR,Li_2018_ECCV}, tried to segment the foreground objects without any given labels. Semi-supervised methods aim to segment the objects in a video with a given ground-truth mask in the first frame. For example, some approaches~\cite{Caelles_2017_CVPR,maninis2018video,voigtlaender2017online,perazzi2017learning} used online fine-tuning to make the model robust to object appearance variation. And some studies~\cite{wug2018fast,johnander2019generative,lin2019agss} based on mask-propagation solely relied on offline training for this task, making the models more efficient. Some~\cite{Zeng_2019_ICCV, sun2020fast} took advantage of Mask R-CNN~\cite{he2017mask} to predict corresponding box of each object and then conduct segmentation. Additionally, Sun et al.~\cite{sun2020adaptive} used reinforcement learning to choose better proposals for target object bounding box and then conduct segmentation. However, offline methods generally are less performing than online ones. In this paper, we will focus on mask-propagation based methods under case of semi-supervision and try to design a fast and high-performance model.

\textbf{Embedding Based Network.} Embedding based networks use an embedding vector to represent each pixel. They have been successfully employed in many vision tasks~\cite{fathi2017semantic,schroff2015facenet,sohn2016improved}. Many successful VOS approaches are also based on embedding. PML~\cite{chen2018blazingly} employed an embedding vector to represent each pixel, and then embedding vectors in reference frame are matched with that of target frame using a triplet loss. VideoMatch~\cite{Hu_2018_ECCV} proposed a matching based algorithm for VOS, which learned to match extracted features to a provided template without memorizing the appearance of target objects. Besides, Ci et al.~\cite{Ci_2018_ECCV} attempted to predict foreground object by learning location-sensitive embedding. FEELVOS~\cite{voigtlaender2019feelvos} proposed a semantic pixel-wise embedding with a global and local matching mechanism for this task, and Yoon et al.~\cite{Yoon_2017_ICCV} utilized features from different depth layers by combinations of convolution, max pooling and Rectified Linear Units to distinguish the target area from the background. However, most embedding based networks need guidance information to tell which pixels belong to foreground and which ones belong to background. In this paper, we directly match the features with our proposed mechanism (directly compute the similarity of pixels) without the separation of positive or negative pools.

\textbf{Cosegmentation.} Some methods utilize cosegmentation to discover video object. VODC~\cite{wang2014video} was proposed to distinguish which frames contained the target object and then segmentation was conducted on these corresponding frames. They designed a spatio-temperal auto-context model to obtain superpixel label for each frame and then a multiple instance boosting algorithm with spatial reasoning was deployed to synchronously detect whether a frame contained the target object and predict the segmentation map. Besides, Wang et al.~\cite{wang2015robust} proposed an energy optimization framework which combined intraframe saliency, interframe consistency and across-video similarity. They used saliency and spatio-temporal SIFT flow to detect initial pixels for common object. Then, the spatio-temporal SIFT was used to refine the coarse object regions generated by the prior step. Additionally, Li et al.~\cite{li2013unsupervised} designed a robust ensemble clustering scheme to predict object-like proposals for unsupervised cosegmentation task. Once the proposals were generated, unary and pairwise energy potentials were minimized with the $\alpha$-expansion to train the model. Although these methods have achieved satisfactory results, they are designed to detect the common object among different video sequences. In our task, we aim to track the same object marked in the first frame for the specific video sequence.

\textbf{Channel Attention Networks.} Channel attention modules have been ever-increasing popular in different computer vision tasks. A multi-channel attention selection mechanism was proposed in SelectionGAN~\cite{tang2019multi} to refine the coarsely generated one on a target image. A residual channel attention network was designed in RCAN~\cite{zhang2018image} to learn the inter-dependencies of features among channels for image super-resolution. SCA-CNN~\cite{chen2017sca} leveraged channel attention to select semantic attributions of corresponding sentence context. Additionally, Qiu et al.~\cite{qiu2020hierarchical} proposed to learn multiple attention maps to obtain hierarchical context information for object detection. All the above methods show that the attention mechanism can help models learn better representations for corresponding targets. Therefore, we consider using the attention module to help our network learn better feature representation for the target object to be tracked and segmented.

\textbf{Non-local Networks.}
Non-local operation is mainly treated as a self-attention mechanism to compute the relationships of the pixels through a global view in the network. Wang et al.~\cite{Wang_2018_CVPR} proposed a non-local operation for capturing long-range dependencies in video classification and static image recognition. DANet~\cite{fu2019dual} plugged non-local operation as position attention module and channel attention module into scene segmentation. In this paper, we introduce the non-local mechanism as a pixel-matching operation to match target pixels and reference pixels to realize the localization of target object in the target frame.

\section{Method}

Our motivation is to make VOS model adaptive to object appearance variation and occlusion, and keep a high efficiency at the same time. Therefore, we design a new mechanism by matching the pixels in target frame and reference frames (first and previous frames) to acquire the predicted mask for the target frame.  

\begin{figure*}
	\centering
		\includegraphics[scale=.4]{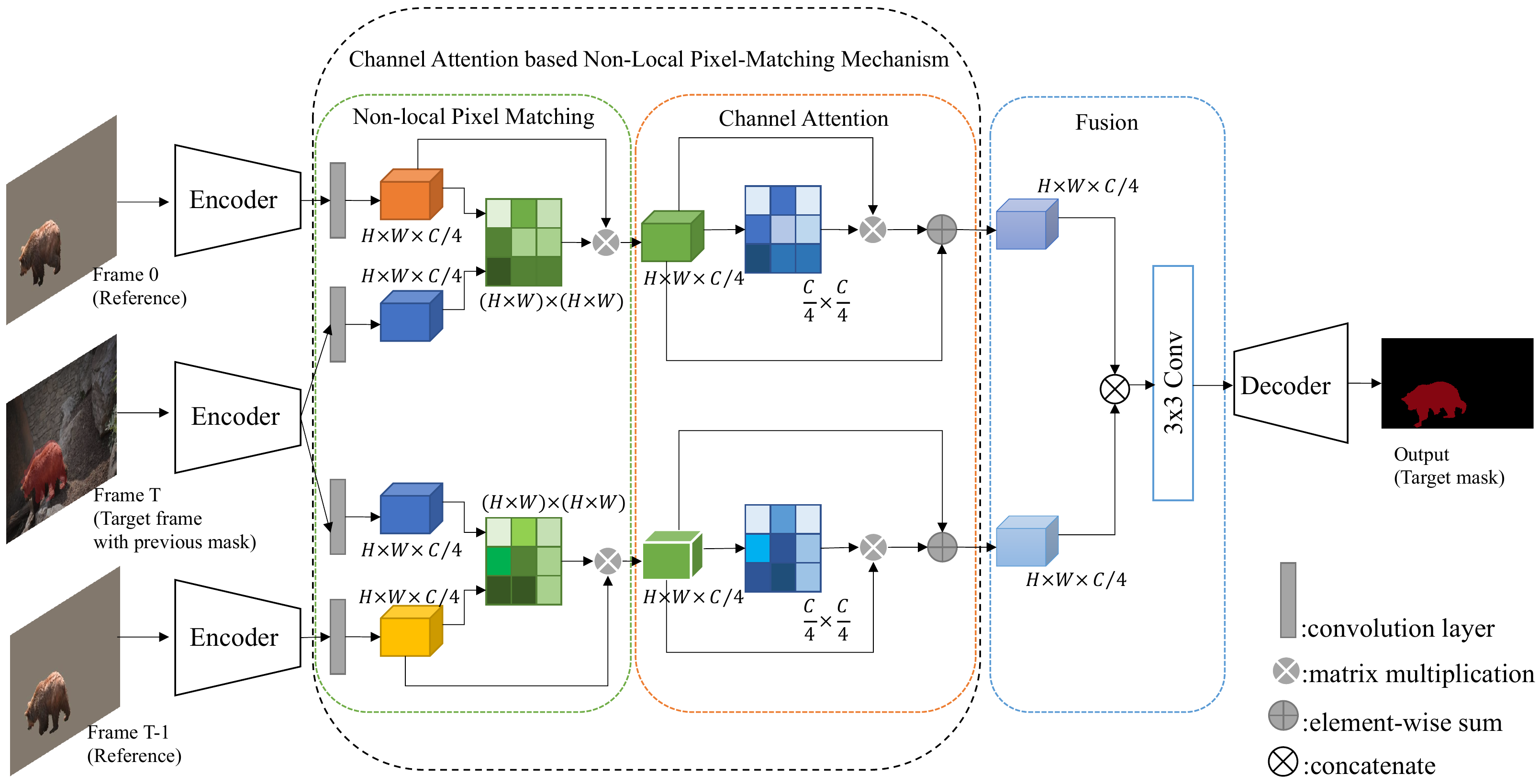}
	\caption{The framework of our NPMCA-net. It consists of three encoders, where the encoders for the two reference frames are shared. NPMCA-net contains a non-local pixel-matching module, a channel attention module, a fusion module and a decoder.}
	\label{FIG:overview}
\end{figure*}

\subsection{Video Object Segmentation Architecture}\label{VOSARCH}
Given a video with annotated mask for the first frame, we need to segment the rest frames according to the given mask. In VOS, object appearance is often changing frame by frame for the video object segmentation task. Thus, it is not sufficient if we only care about the object appearance in the first frame, especially when large object appearance variation occurs in the middle of the video. 

As illustrated in Fig.~\ref{FIG:overview}, we provide three different kinds data for the three encoders: the target frame encoder takes the current frame with the estimated labels of the previous frame as 4-channel input~\cite{wug2018fast}; two parameter-shared reference frame encoders take the first frame and the previous frame as input, respectively. Note that when providing data for reference frame encoders, background pixels from the first frame and the previous frame are removed using groundtruth (first frame) or estimated mask (previous frame). Whist for the target frame encoder, background pixels are not masked-out since the masks for the current and previous frames are different. Then, the feature maps of reference and target frames are extracted by respective encoders. In this way, we can obtain the changing object appearance information and target frame features. 

Following that, the feature maps are input into our non-local pixel-matching module. The target feature map is matched with the feature maps from two references using our newly designed non-local pixel-matching module to localize the target objects. In this process, the target feature is matched with two references one by one, individually. Therefore, there are two output feature maps: one is the matched feature map of the target frame with the first frame, and the other one is the matched feature map of target with previous frame. With the help of the previous frame, our network can adapt to object appearance variation, since the gap between the current and previous frames are smaller than that between the current and first frame. On the other hand, if we only consider the previous frame, for the occlusion case, the model will lose the initial object appearance for frames after the occlusion.

After that, the channel attention module is applied to strengthen features by allocating different weights for each feature channel. Once the features are matched and enhanced, the obtained two feature maps are concatenated, where a $3\times3$ convolution layer is used to fuse the two feature maps. Finally, the fused feature map is decoded by the decoder to predict and output the target object masks. Our method can be viewed as an encoder-decoder process, which can directly obtain the segmentation mask of current frame without any post-processing.

\begin{figure}
	\centering
	\subfigure[The Process of Similarity Computation]{
		\label{FIG:similarity}
		\includegraphics[width=3.5in]{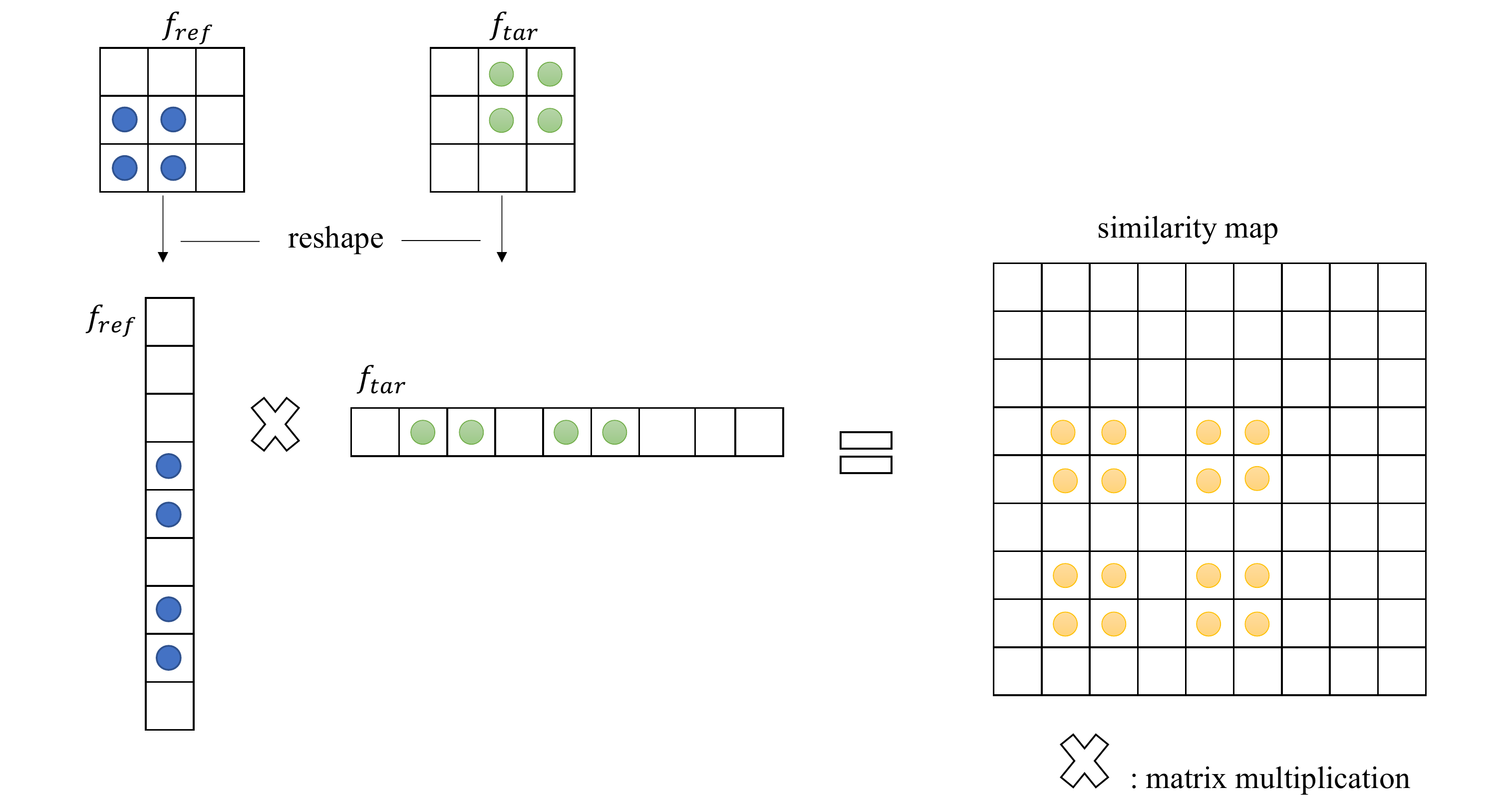}
	}
	\hspace{1in}
	\subfigure[The Process of Matching and Localization]{
		\label{FIG:matching}
		\includegraphics[width=4.0in]{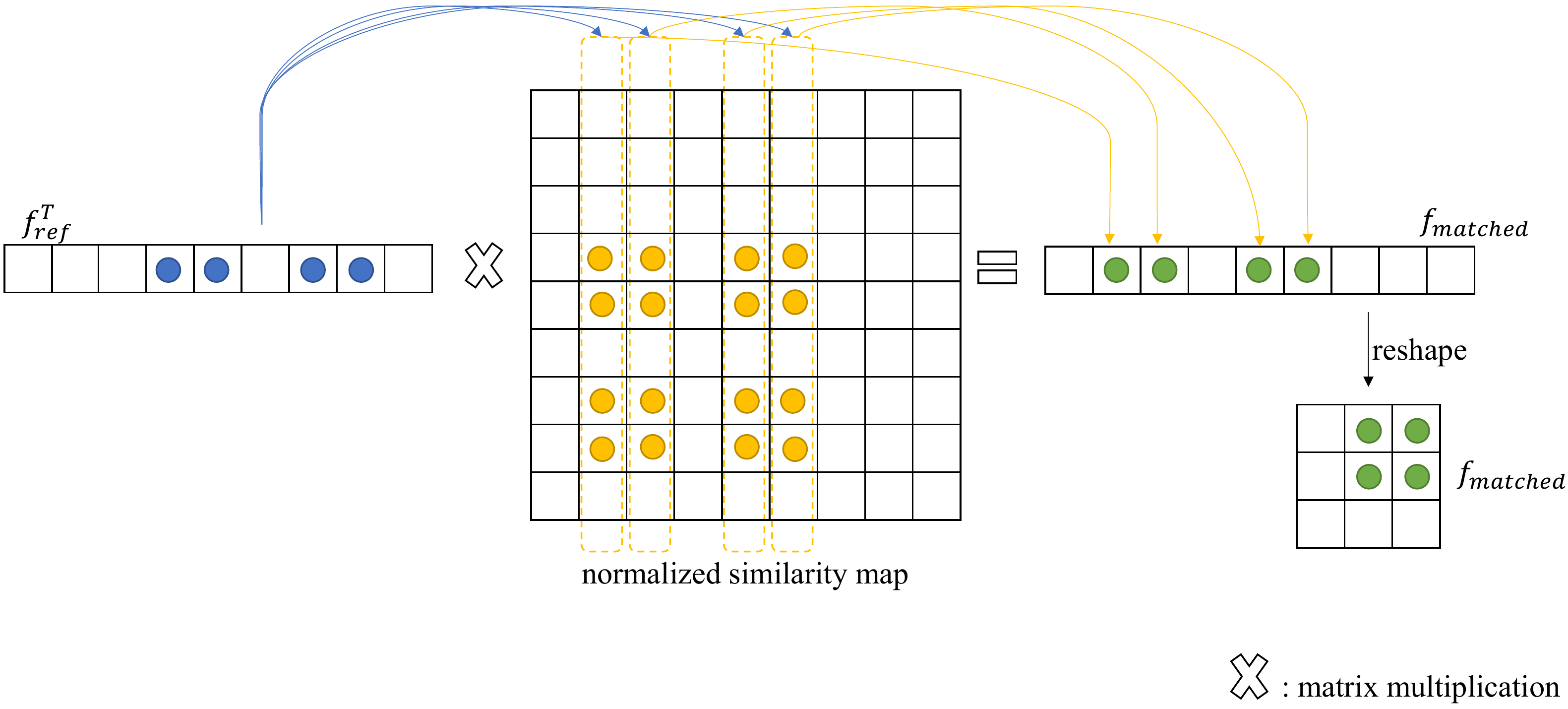}
	}
	\caption{(a) The process of similarity computation (Eq.(\ref{eq:similarity})). The two reduced feature maps are reshaped into $f_{ref}\in \mathbb{R}^{N\times \frac{C}{4}}$ and  $f_{tar}\in \mathbb{R}^{ \frac{C}{4} \times N}$, and the similarity is computed by the matrix multiplication. (b) The process of target object matching and localization (Eq.(\ref{eq:fmatched})). }
	\label{FIG:NLPMM process}
\end{figure}

\begin{figure}
    \centering
    \subfigure[Non-Local Pixel-Matching Module]{
        \label{FIG:nlpmm}
        \includegraphics[width=4.0in]{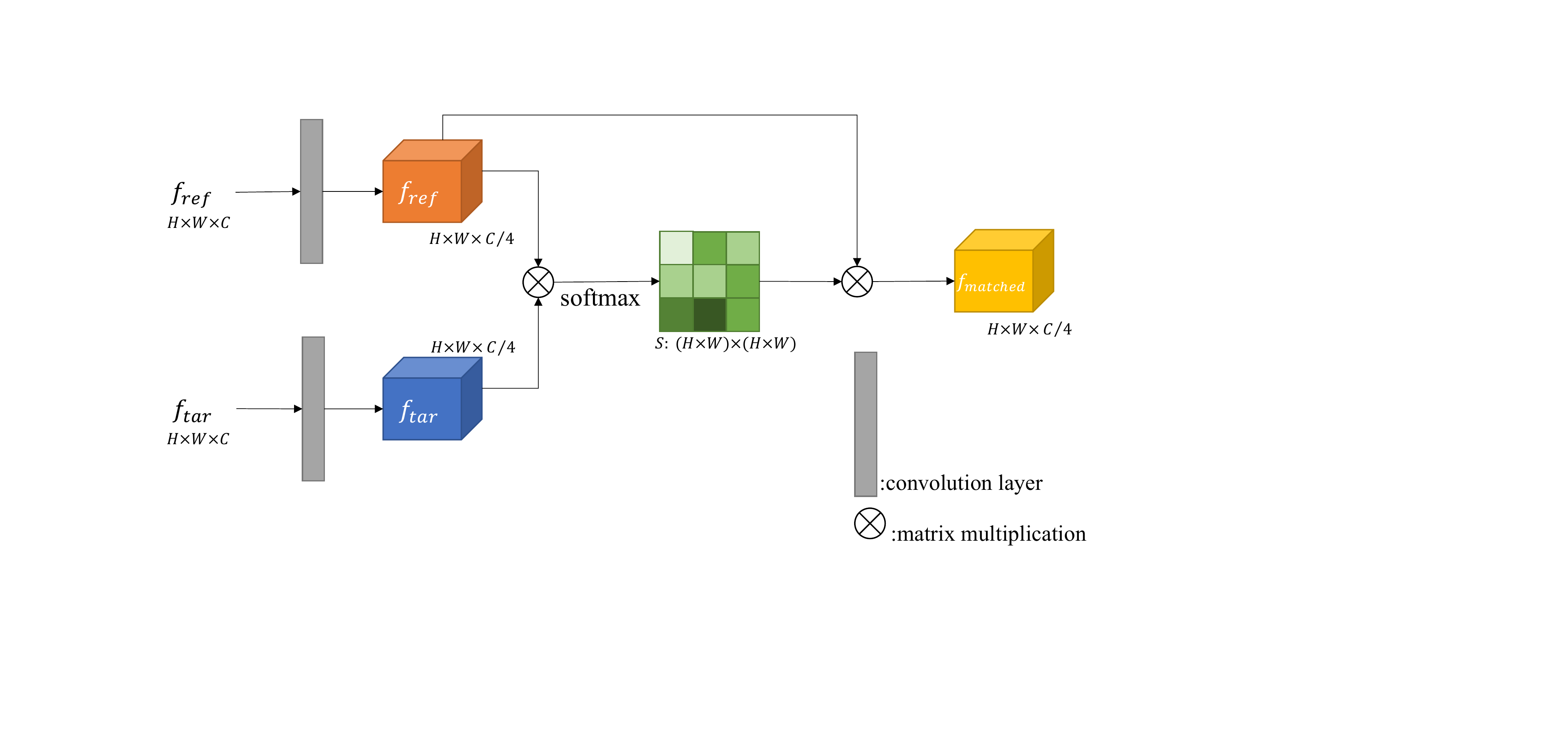}
    }
    \hspace{1in}
    \subfigure[Visualization of Output Feature Map of NLPMM]{
        \label{FIG:nlpmmoutput}
        \includegraphics[width=3.5in]{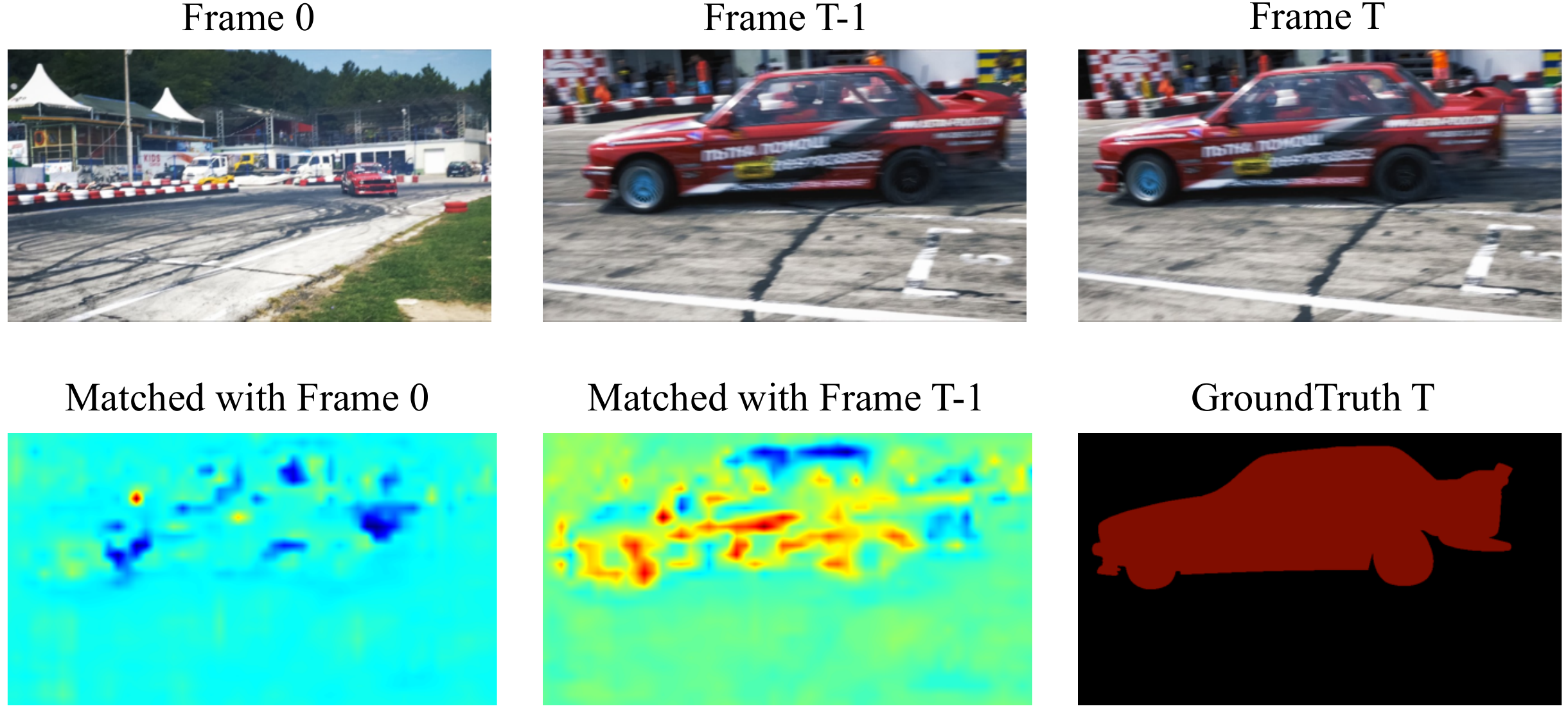}
    }
    \caption{(a) Framework of non-local pixel-matching module (NLPMM). Our NLPMM has two inputs,including the reference feature map and the target feature map. The output is the matched feature map. (b) Visualization of output feature map from NLPMM. The matched feature map can coarsely acquire the foreground object appearance and its location.}
    \label{FIG:moduleframework}
\end{figure}

\subsection{Non-Local Pixel Matching with Channel Attention}\label{NPMCA-net}

Our NPMCA-net contains two parts, including a non-local pixel-matching module (NLPMM) and a channel attention module (CM). The CM is in series with the NLPMM. The NLPMM is a non-local structure which can match pixels over the whole feature map. And CM conducts self-attention through the channel dimension instead of the spatial dimension to strengthen the feature representation. With the combination of these two modules, our network can obtain feature representations of the foreground objects for the target frame. The details are discussed as follows.

\textbf{Non-Local Pixel-Matching Module.}
The non-local pixel-matching module is one main module of our NPMCA-net, which is used to obtain object appearance of the target frame and localize the target object simultaneously by matching the feature maps of the reference frames and the target frame. Different from the matching process using convolution layers~\cite{Yoon_2017_ICCV} or  using metric learning to pull in similar embedding vectors and push away different embedding vectors~\cite{voigtlaender2019feelvos,chen2018blazingly}, we directly compute similarities between pixels. The framework of NLPMM is illustrated in Fig.\ref{FIG:nlpmm}. The inputs of this module are the feature map of reference frame and the feature map of target frame (defined as $f_{ref}\in \mathbb{R}^{H\times W\times C}$ and $f_{tar}\in \mathbb{R}^{H\times W\times C}$, where $H, W, C$ are the height, width, and channel number, respectively) extracted from respective encoders. In order to reduce memory and improve efficiency for our approach, once feature maps are fed into the module, a $3\times3$ convolution layer with padding is used to reduce the channel number of input feature maps from $C$ to $C/4$, the new feature maps are with size $f_{ref}\in \mathbb{R}^{H\times W\times \frac{C}{4}}$ and  $f_{tar}\in \mathbb{R}^{H\times W\times \frac{C}{4}}$, respectively. After that, the two reduced feature maps are reshaped to $f_{ref}\in \mathbb{R}^{N\times \frac{C}{4}}$ and  $f_{tar}\in \mathbb{R}^{N\times \frac{C}{4}}$,  where $N = H \times W$. The similarity between pixels in the two feature maps is computed: 
\begin{equation}
    \centering
    S = f_{ref}  f_{tar}^T,
    \label{eq:similarity}
\end{equation}
with $S(i,j)$ measuring the similarity between $i^{th}$ position on reference feature map and $j^{th}$ position on target feature map. The similarity of each pixel is calculated in a non-local way, where all positions of the two feature maps are included. Meanwhile, it computes the relation between two spatial pixels from two temporal frames because the inputs are from a temporal sequence. Therefore, it is a space-temporal similarity calculation. After that, instead of directly using the calculated result, we apply softmax to normalize the non-local similarity map $S$, and obtain $S'$ ($S' \in \mathbb{R}^{N\times N}$, $N=H \times W$), with its element value $S'(i,j)$ being 
\begin{equation}
    \centering
    S'(i,j) = \frac{exp(S(i,j))}{\sum_{i = 1}^{N}exp(S(i,j))}.
    \label{eq:norm}
\end{equation}
With Eq.(\ref{eq:similarity}) and Eq.(\ref{eq:norm}), we can generate the relations between any two pixels in the target feature map and the reference feature map. The pixel pair with a large similarity value has high probability belonging to the same pixel of one foreground object. In this case, we can not only match the object appearance but also localize the object. Finally, the new matched feature map $f_{matched}$ is calculated by a matrix multiplication between the transpose of the reduced reference feature map $f_{ref}$ and the non-local similarity map $S'$ ,
\begin{equation}
    f_{matched} = f_{ref}^T S'
    \label{eq:fmatched}.
\end{equation}
Finally, the matched feature map is reshaped back to $f_{matched}\in \mathbb{R}^{H\times W\times \frac{C}{4}}$.

The coarse mask of the target frame can be obtained by the matrix multiplication between the reference feature map and the similarity map, namely, we can use Eq.(\ref{eq:fmatched}) to obtain the pixels of foreground objects in the target frame.  To more intuitively understand the matching and localization process, we show the process in Fig.\ref{FIG:NLPMM process}. Fig.\ref{FIG:similarity} shows how the similarity map is computed, and Fig.\ref{FIG:matching} displays how the matching process can also accomplish the localization. Therefore, we can obtain foreground object appearance and its location at the same time. Besides, visualization of the output of our non-local pixel-matching module is shown in Fig.\ref{FIG:nlpmmoutput}. It can be found that this matching module is able to localize the object and mask the target object appearance. The highlighted part (warm color) in the “matched with frame T-1” better demonstrates the matched pixels for the target object. When there is only frame 0 to be referred, it is difficult for the network to find out the pixels for the moving object in the case of large appearance variation.

\textbf{Channel Attention Module.}
\begin{figure}
    \centering
    \subfigure[Channel Attention Module]{
        \label{FIG:cam}
        \includegraphics[width=4.0in]{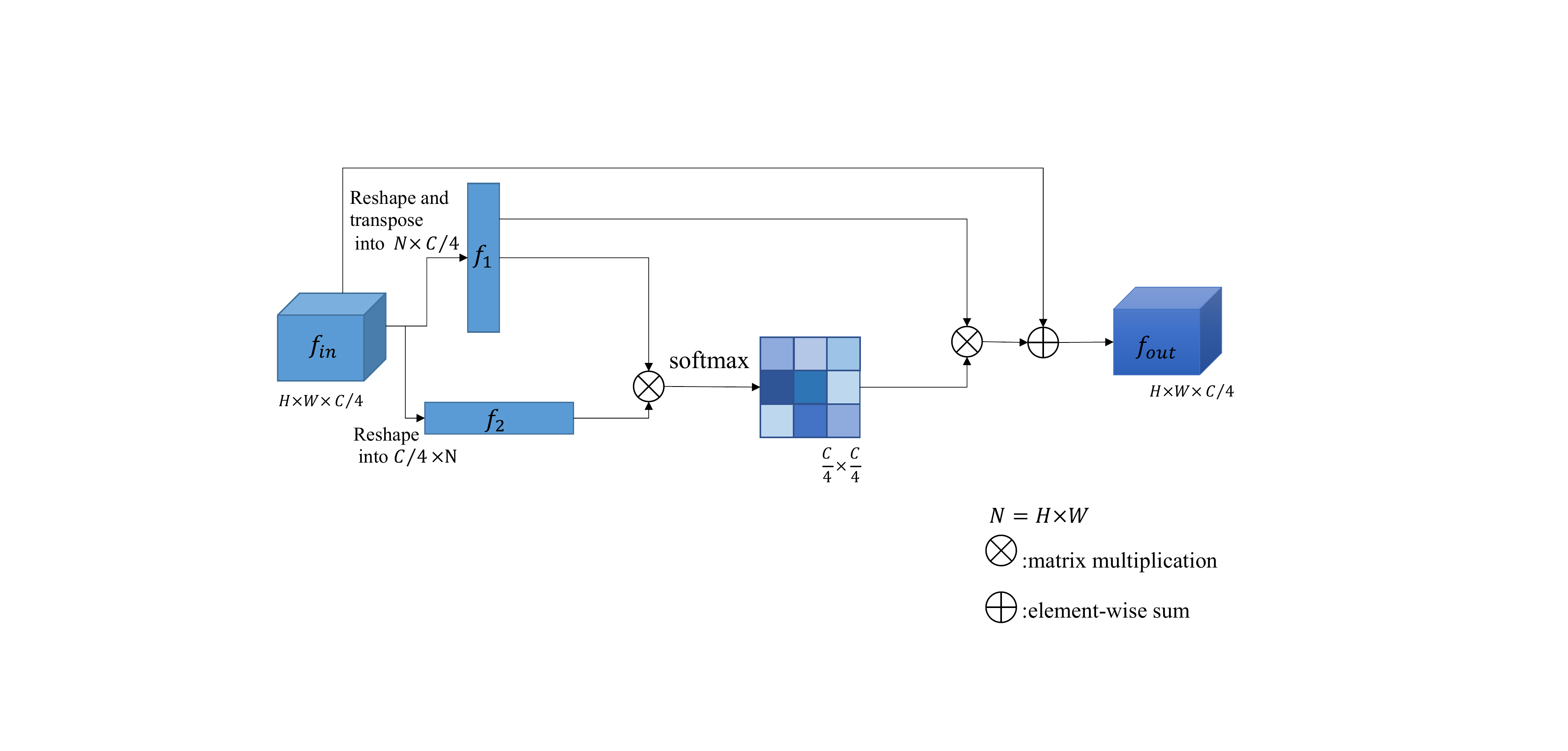}
    }
    \hspace{1in}
    \subfigure[Visualization of Output Feature Map of CM]{
        \label{FIG:camoutput}
        \includegraphics[width=3.5in]{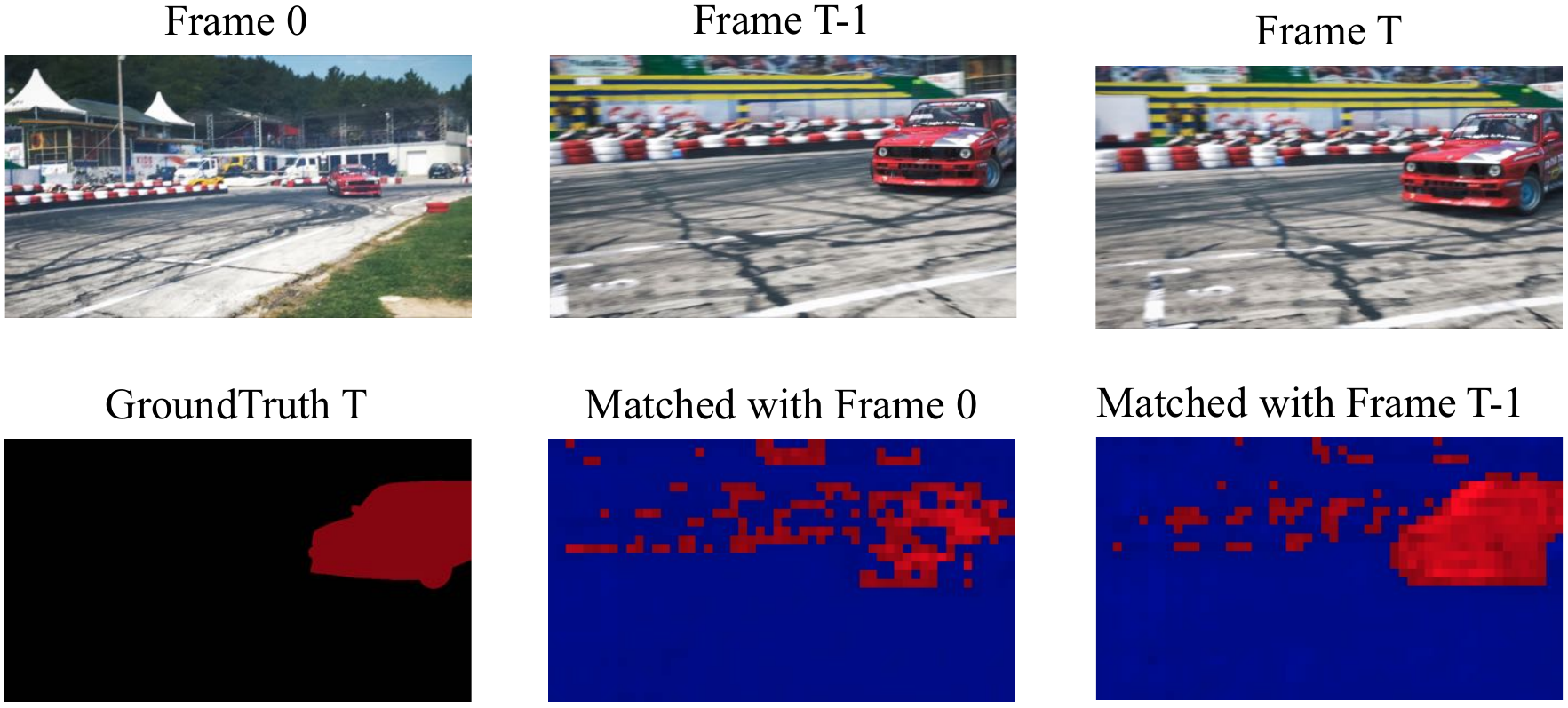}
    }
    \caption{(a) Framework of channel attention module (CM). The input of CM is the output of NLPMM (matched feature map), and it outputs the strengthened feature map. (b) Visualization of Output feature map from CM. CM is able to strengthen the feature representation.}
    \label{FIG:cammoduleframework}
\end{figure}
We adopt a channel attention module after the non-local pixel-matching module to strengthen the feature representation of foreground object in this task. The details of our channel attention module is illustrated in Fig.\ref{FIG:cam}. The input for this module $f_{in}$ is the output feature map of non-local pixel-matching module, \emph{i.e.}, $f_{in} = f_{matched}$ and $f_{in} \in \mathbb{R}^{H\times W\times \frac{C}{4}}$. In order to compute the inter-dependencies between different channels, $f_{in}$ is first reshaped into $f_{in}\in \mathbb{R}^{N\times \frac{C}{4}}$, where $N=H\times W$. Then the channel attention map $A\in \mathbb{R}^{\frac{C}{4}\times \frac{C}{4}}$ is computed by:
\begin{equation}
    \centering
    A = f^T_{in} f_{in},
    \label{eq:ca}
\end{equation}
\begin{equation}
    \centering
    A'(i,j) = \frac{exp(A(i,j))}{\sum_{i = 1}^{N}exp(A(i,j))}
    \label{eq:ca},
\end{equation}
where $A(i,j)$ measures the relationship between $i^{th}$ channel and $j^{th}$ channel of $f_{in}$. Then matrix multiplication is applied to get the strengthened feature map. Mathematically, the strengthened feature is:
\begin{equation}
    f_{A} = f_{in} A'.
\end{equation}

Then the strengthened feature map $f_{A}$ is reshaped back into the size of input feature map, \emph{i.e.}, $f_{A}\in \mathbb{R}^{H\times W\times \frac{C}{4}}$. The final output of channel attention module is the weighted sum of the strengthened feature map and the module input feature map $f_{in}$:
\begin{equation}
    f_{out} =  \gamma f_A + f_{in},
\end{equation}
where $\gamma \geq 0$ is a learned parameter. We do not apply any convolution layer in the channel attention map. The channel attention map is in series with the non-local pixel-matching module to strengthen the representation of feature map instead of adopting a parallel mode in~\cite{fu2019dual}. Some visualizations of the output feature map of the channel attention module are displayed in Fig.\ref{FIG:camoutput}.

\subsection{Two-stage Training Method}
We take two-stage training for our network. Firstly, we pre-train our NPMCA-net through static images. Then, we use the video object segmentation datasets to fine-tune the model. We use IoU loss in~\cite{lin2019agss,li2018interactive} and Adam~\cite{kingma2014adam} optimizer with randomly cropped resolution of ($256 \times 432$) patches for both pre-training and fine-tuning. All experiments are running on one NVIDIA GeForce 2080 Ti GPU.

\textbf{Pre-training on static images.}
Pre-training on static images for video object segmentation is becoming popular recently since it can help the network adapt to different foreground object appearance. We follow several successful practice in ~\cite{wug2018fast,perazzi2017learning,oh2019video} to pre-train our network by applying random affine transformation on static images. We use saliency datasets MSRA10K~\cite{cheng2014global}, ECSSD~\cite{yan2013hierarchical}, segmentation datasets Pascal VOC dataset~\cite{everingham2010pascal} and COCO~\cite{lin2014microsoft}. In this case, the network can be adapted to different object appearance and categories, so as to avoid easy over-fitting. For pre-training, we set a fixed learning rate as 1$e$-5.

\textbf{Fine-tuning on videos.}
Then, we fine-tune the pre-trained model on video object segmentation dataset. We only use DAVIS-17~\cite{pont20172017} training set for fine-tuning. During training, we sample three frames in temporal order to obtain temporal information. In order to acquire big variation of object appearance for a long time, we randomly skip frames for sampling. The maximum random skip is 5 and the learning rate  for fine-tuning is set as 1$e$-6.

\section{Experiment}
\subsection{Inference}
Our network is based on the assumption that the ground-truth mask of the first frame is given for semi-supervised video object segmentation. In other words, the first frame is set as the reference frame for all the rest frames. Therefore, to make our network efficient, we only compute the feature map of the first frame once for a test video clip. Following the architecture of our approach, we use previous frame with predicted segmentation mask as another reference frame. We also follow \cite{wug2018fast} to set three different scale sizes and compute their average as the final output.

\textbf{Multi-object case.}
We use softmax aggregation~\cite{wug2018fast} to softly combine multiple objects. Finally, the output probability map is computed by:
\begin{equation}
    P_{i,m}=\frac{p_{i,m}/(1-p_{i,m})}{\sum_{j = 0}^{M}p_{i,j}/(1-p_{i,j})},
    \label{eq:mul}
\end{equation}
where $p_{i,m}$ is the output probability of instance $m$ at position $i$. $m=0$ is for background and $M$ is the total number of instances. We use Eq.(\ref{eq:mul}) to compute the probability map of multi-objects and apply it to next frame inference.

\subsection{Implementation}
 \textbf{Encoder.}  We design three encoders based on ResNet-50~\cite{he2016deep} for three inputs (two references and one target). Like~\cite{wug2018fast}, the target frame encoder takes 4-channel inputs and two reference frame encoders take 3-channel inputs. Instead of using res5 in~\cite{wug2018fast}, we take res4 as the final encoded feature map, whose channel number is 1024. This is because the feature map of res5 is with low resolution, making it inaccurate for small objects. On the other hand, three res5 encoders will cause large memory occupation.
 
 \textbf{Decoder.} After the fusion layer, the fused feature map is finally fed into the decoder. Similar to~\cite{wug2018fast}, the decoder also takes the encoder stream through skip-connection as input to produce the mask. With the help of skip-connection, the high resolution feature can replenish the missing information. Finally, the feature map is gradually upsampled with a factor of two till it reaches the same size as input.

\subsection{Experiment Results}
We evaluate our network on video object segmentation datasets, DAVIS-2017~\cite{pont20172017}, DAVIS-2016~\cite{perazzi2016benchmark} and SegTrack-v2~\cite{FliICCV2013}. The evaluation metrics include mean intersection-over-union (IoU) of predicted mask and the ground-truth ($\mathcal{J}$), contour accuracy between contour points on predicted mask and the ground-truth ($\mathcal{F}$), and the average of the two metrics ($\mathcal{J}$\&$\mathcal{F}$).

\textbf{DAVIS-2017.} DAVIS-2017 is a multi-object dataset. There are 90 videos in total, 60 for training and 30 for validation. We evaluate our method on its validation set. The comparison results with recent state-of-the-art approaches are shown in Table~\ref{tbld17}. The results are listed from the lowest score of $\mathcal{J}$ to the highest score. The upper part is from approaches with online-learning or with optical flow. It can be found that our method achieves comparable scores with the best performing ones. Our score is slightly lower than PReMVOS~\cite{luiten2018premvos}, but PReMVOS needs longer running time than all other approaches because both online-learning and optical flow need expensive computational cost. We reach the best performance compared with all other methods without online-learning or optical flow. It can be demonstrated that our NLPMM can realize find out where the target object is in current frame. Further, we directly using masked-out object as the input for reference, making our model less sensitive to the influence of backgrounds while focusing on the object itself. By doing this, our method can capture enough object features. Besides, using the masked-out objects of the first frame and the previous frame as references provides enough information for handling appearance variation.

\begin{table}[width=.9\linewidth,cols=6,pos=h]
\caption{Evaluation on DAVIS-17 validation set. `OL' denotes online-learning. `OF' means using optical flow. Our NPMCA-net obtains a score of 3$\%$ higher than STM~\cite{oh2019video}.}\label{tbld17}
\centering
\begin{tabular}{c|cc|ccc|c}
\toprule
Method & OL & OF & $\mathcal{J}$ ($\%$) & $\mathcal{F}$ ($\%$) & $\mathcal{J}$\&$\mathcal{F}$ ($\%$) & Time (s) \\
\hline
OSVOS~\cite{Caelles_2017_CVPR} & \checkmark &  & 56.6 & 63.9 & 60.3 & 10\\
OnAVOS~\cite{voigtlaender2017online} & \checkmark &  & 61.6 & 69.1 & 65.4 & 13 \\
OSVOS-S~\cite{maninis2018video} & \checkmark &  & 64.7 & 71.3 & 68.0 & 4.5 \\
AGSS-VOS~\cite{lin2019agss} &  & \checkmark & 64.9 & 69.9 & 67.4 & - \\
CINN~\cite{Bao_2018_CVPR} & \checkmark &  & 67.2 & 74.2 & 70.7 & >120\\
PReMVOS\cite{luiten2018premvos} & \checkmark & \checkmark & \textbf{73.9} & \textbf{81.8} & \textbf{77.8} & - \\
\hline
VideoMatch\cite{Hu_2018_ECCV} &  &  & 56.5 & 68.2 & 62.4  & 0.35\\
MAARU~\cite{fu2021video} & & & 61.3 & 65.3 & 63.3 & 0.13\\
RANet~\cite{wang2019ranet} &  &  & 63.2 & 68.2 & 65.7 &-\\
RGMP~\cite{wug2018fast} &  &  & 64.8 & 68.6 & 66.7  & 0.28\\
DIPNet~\cite{hu2020dipnet} & & & 65.3 & 71.6 & 68.5 & -\\
A-GAME~\cite{johnander2019generative} &  &  & 67.2 & 72.7 & 70.0&- \\
DMM-Net~\cite{Zeng_2019_ICCV} &  &  & 68.1 & 73.3 & 70.7&- \\
FEELVOS~\cite{voigtlaender2019feelvos} &  &  & 69.1 & 74.0 & 71.6 & 0.51 \\
STM~\cite{oh2019video} &  &  & 69.2 & 74.0 & 71.6 &-\\
TVOS~\cite{zhang2020transductive} & & & 69.9 & 74.7 & 72.3 & 0.027\\
NPMCA-net (Ours) &  &  & \textbf{72.2} & \textbf{77.4} & \textbf{74.8} & 0.25\\
\bottomrule
\end{tabular}
\end{table}

\textbf{DAVIS-2016.} DAVIS-2016 contains 50 videos (30 for training and 20 for validation) for single-object video object segmentation. We report comparison results of the validation set in Table~\ref{tbld16}. It can be found that our approach achieves better performance than the methods using pixel-matching or metric learning, such as PLM~\cite{Yoon_2017_ICCV}, PML~\cite{chen2018blazingly}, FEELVOS~\cite{voigtlaender2019feelvos}, and RGMP~\cite{wug2018fast}. We also obtain higher score than other methods without online learning. For metric $\mathcal{J}$, our method is 1.7$\%$ higher than STM~\cite{oh2019video}, whist for the contour accuracy, our method is 0.8$\%$ lower than STM~\cite{oh2019video}, this might be caused by the adopted IoU loss. Moreover, our results are competitive with online-learning based methods. According to the running time listed in Table~\ref{tbld16}, our approach can achieve a good balance between accuracy and efficiency. It demonstrates that our NLPMM is able to localize moving objects with masked-out object references. Additionally, pre-training with statistic images also helps network to adapt to different object classes. In this way, our approach does not rely on online training to learn the object information of current video.

\begin{table}[width=.9\linewidth,cols=7,pos=h]
\caption{Evaluation on DAVIS-16 validation set. `OL' denotes online-learning. `OF' means using optical flow. Our NPMCA-net can even achieve a bit higher performance than methods with online-learning.}\label{tbld16}
\begin{tabular}{c|cc|ccc|c}
\toprule
Method & OL & OF & $\mathcal{J}$ ($\%$) & $\mathcal{F}$ ($\%$) & $\mathcal{J}$\&$\mathcal{F}$ ($\%$) & Time (s) \\
\hline
MSK~\cite{perazzi2017learning} & \checkmark & \checkmark & 79.7 & 75.4 & 77.6 & 12 \\
OSVOS~\cite{Caelles_2017_CVPR} & \checkmark &  & 79.8 & 80.6 & 80.2 & 7 \\
MaskRNN~\cite{hu2017maskrnn} & \checkmark & \checkmark & 80.7 & 80.9 & 80.8 & - \\
CINN~\cite{Bao_2018_CVPR} & \checkmark &  & 83.4 & 85.0 & 84.2 & \textgreater30 \\
Lucid~\cite{khoreva2017lucid} & \checkmark & \checkmark & 83.9 & 82.0 & 83.0 & - \\
PReMVOS~\cite{luiten2018premvos} & \checkmark & \checkmark & 84.9 & \textbf{88.6} & \textbf{86.8} & \textgreater30 \\
OSVOS-S~\cite{maninis2018video} & \checkmark &  & 85.6 & 86.4 & 86.0 & 4.5 \\
OnAVOS~\cite{voigtlaender2017online} & \checkmark &  & 86.1 & 84.9 & 85.5 & 13 \\
DyeNet~\cite{li2018video} & \checkmark &  & \textbf{86.2} & - & - & 2.32 \\
\hline
PLM~\cite{Yoon_2017_ICCV} &  &  & 70.0 & 62.0 & 66.0 & 0.3 \\
PML~\cite{chen2018blazingly} &  &  & 75.5 & 79.3 & 77.4 & 0.28 \\
VideoMatch\cite{Hu_2018_ECCV} &  &  & 81.0 & - & - & 0.32 \\
FEELVOS~\cite{voigtlaender2019feelvos} &  &  & 81.1 & 82.2 & 81.7 & 0.45 \\
RGMP~\cite{wug2018fast} &  &  & 81.5 & 82.0 & 81.8 & 0.13 \\
A-GAME~\cite{johnander2019generative} &  &  & 82.0 & 82.2 & 82.1 & 0.07 \\
MAARU~\cite{fu2021video} & & & 83.9 & 83.8 & 83.9 & 0.12\\
RANet~\cite{wang2019ranet} &  &  & 85.5 & 85.4 & 85.5 & 0.13 \\
DIPNet~\cite{hu2020dipnet} & & & 85.8 & 86.4 & 86.1 & 0.92\\
STM~\cite{oh2019video} &  &  & 84.8 & \textbf{88.1} & 86.5 & 0.15 \\
NPMCA-net (Ours) &  &  & \textbf{86.5} & 87.3 & \textbf{86.9} & 0.11 \\
\bottomrule
\end{tabular}
\end{table}

\textbf{SegTrack v2.} We also evaluate our network on the SegTrack v2~\cite{FliICCV2013} dataset. The results are shown in Table~\ref{tblsegtrack}. It can be found that our network also achieve competitive performance on SegTrack v2 dataset under the same level comparison. Therefore, our network has competitive generalization ability. Our performance even defeat MSK~\cite{perazzi2017learning} and MaskRNN~\cite{hu2017maskrnn}, where online training is used. We set the same training dataset as DMM-net. it can be seen that our method can obtain comparable results with DMM-net. However, we obtain lower performance than DyeNet. This phenomenon may be caused by the fact that they use template matching, which predicts bounding box of the target object first then conduct segmentation. In this way, much background noise can be reduced. In the SegTrack v2 dataset, there are several videos with the background very similar to the target object. In such cases, template can better decrease the disturbance of background. However, for other datasets, such as, DAVIS17, DAVIS16, such conditions are not satisfied, the performance of DyeNet is lower than ours, as reported in Table~\ref{tbld17} and Table~\ref{tbld16}.

\begin{table}[width=.9\linewidth,cols=4,pos=h]
\caption{Evaluation on SegTrack v2. The IoU peformance for the baseline methods are from  ~\cite{wug2018fast} and~\cite{Zeng_2019_ICCV}. `OL' denotes online-learning.}\label{tblsegtrack}
\centering
\begin{tabular}{c|c|c}
\toprule
Method & OL & IoU ($\%$) \\
\hline
OnAVOS~\cite{voigtlaender2017online} & \checkmark &  66.7 \\
MSK~\cite{perazzi2017learning} & \checkmark & 70.3 \\
MaskRNN~\cite{hu2017maskrnn} & \checkmark & 72.1 \\
CINN~\cite{Bao_2018_CVPR} & \checkmark &  77.1 \\
Lucid~\cite{khoreva2017lucid} & \checkmark & \textbf{77.6} \\
\hline
RGMP~\cite{wug2018fast} &  & 71.1 \\
DIPNet~\cite{hu2020dipnet} & & 73.8\\
DMM-Net~\cite{Zeng_2019_ICCV} &  & 76.7 \\
DyeNet~\cite{li2018video} & & 78.3 \\
\hline
NPMCA-net (Ours) &  & 76.1 \\
\bottomrule
\end{tabular}
\end{table}

\subsection{Qualitative Results}
\begin{figure*}
    \centering
    \subfigure[The visual results of our NPMCA-net on DAVIS-2016.]{
        \label{FIG:16results}
        \includegraphics[width=5.0in]{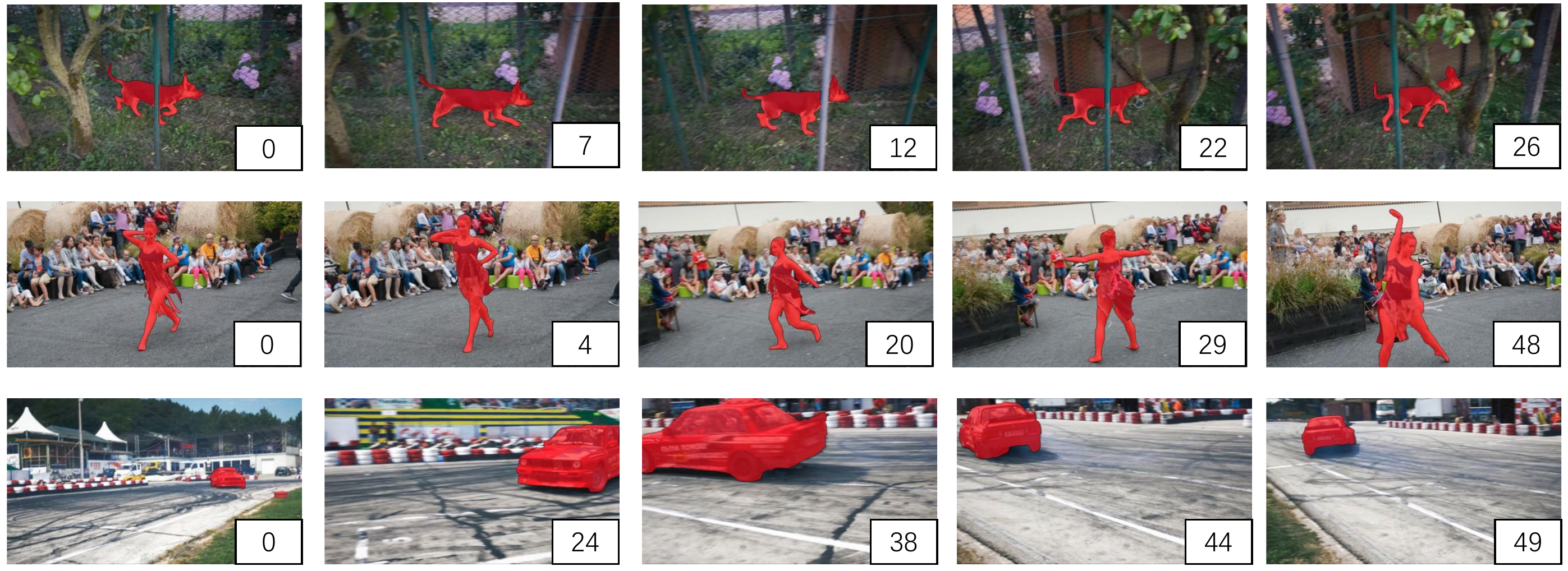}
    }
    \hspace{1in}
    \subfigure[The visual results of our NPMCA-net on DAVIS-2017.]{
        \label{FIG:17results}
        \includegraphics[width=5.0in]{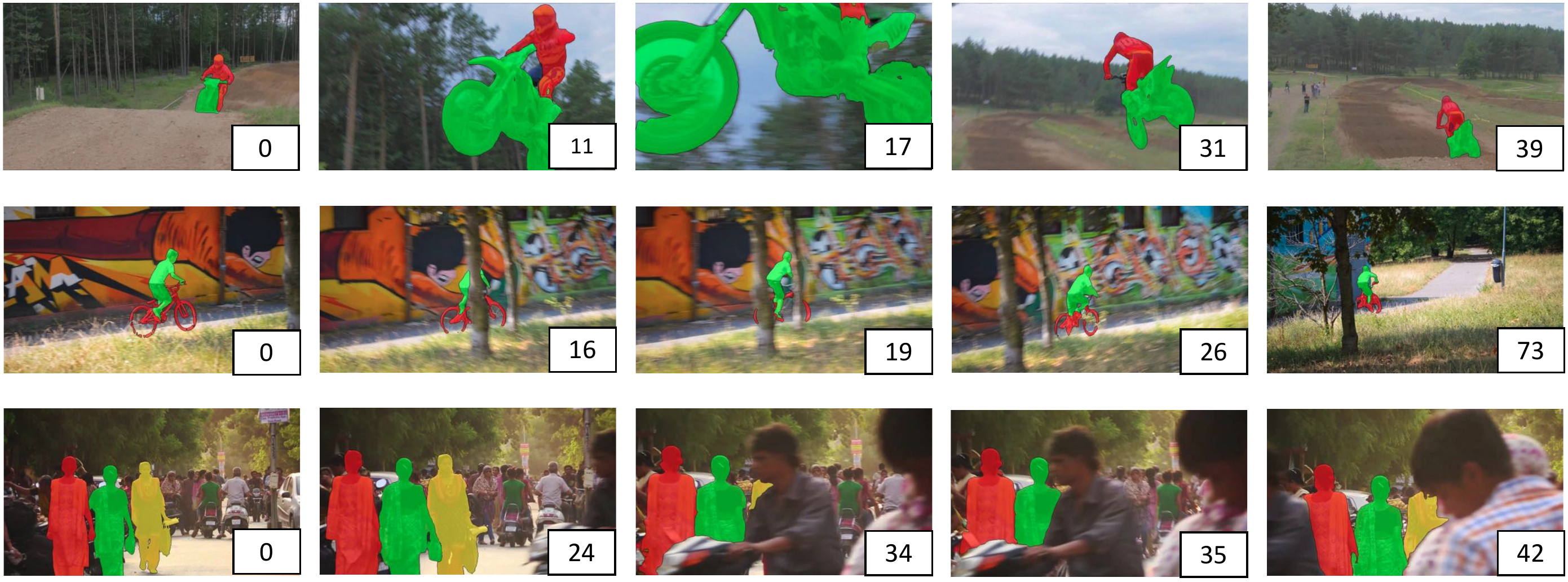}
    }
    \hspace{1in}
    \subfigure[The visual comparison with other approaches on DAVIS-2017.]{
        \label{FIG:comparison}
        \includegraphics[width=5.0in]{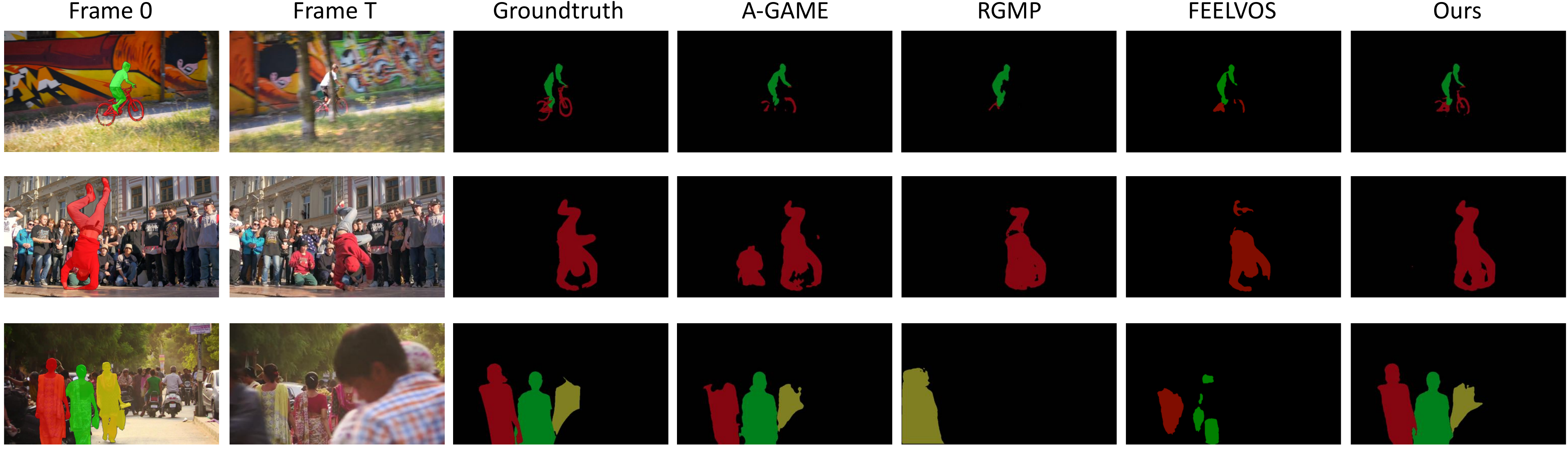}
    }
    \caption{We display the frames with large appearance variation or before and after occlusion and the comparison between ours and other approaches.}
    \label{FIG:visualresults}
\end{figure*}

Qualitative results on two DAVIS datasets are shown in Fig.~\ref{FIG:visualresults}. For each displayed video, we choose 5 frames with the cases of large object appearance variation or occlusion. It can be found that our model can handle different challenges. For example, our model performs well with large object appearance variation cases like in row 2 and 3 in Fig.~\ref{FIG:16results} and row 1 in Fig.~\ref{FIG:17results}. Besides, our model can also segment each object when they are occluded by background as shown in row 1 in Fig.~\ref{FIG:16results} and row 2, 3 in Fig.~\ref{FIG:17results}. The qualitative comparison between our model and other methods are shown in Fig.~\ref{FIG:comparison}.

\subsection{Ablation Studies}

\textbf{Two-stage training method.}
We firstly conduct the ablation study for the two-stage training method, and the results are displayed in Table~\ref{tbltst}. It is surprising to find that the performance of pre-train-only case is much better than fine-tune-only case. Both the intersection-over-union score ($\mathcal{J}$) and the contour accuracy ($\mathcal{F}$) of pre-train-only are almost 25$\%$ larger than of fine-tuning-only. It proves that two-stage training is necessary. If we only train on DAVIS-2017, the categories are far less enough. It can also be found that our approach will perform better when more categories are used for training. The combination of pre-train and fine-tuning achieves the best performance, because pre-training help our model adapt to large categories and fine-tuning help our model to obtain temporal information and adapt to video sequence.

\begin{table}[width=.9\linewidth,cols=3,pos=h]
\caption{Training methods analysis on DAVIS-2017 validation set. The two-stage training method helps our NPMCA-net better adapt to different categories. With only DAVIS-2017 training set, the network is easy to get over-fitting.}\label{tbltst}
\begin{tabular}{c|cc}
\toprule
Training Method & $\mathcal{J} ($\%$)$ & $\mathcal{F} ($\%$)$ \\
\hline
Pre-train only & 65.7 & 71.3 \\
Fine-tuning only & 41.0 & 43.9 \\
\hline
Full Training & \textbf{72.2} & \textbf{77.4} \\
\bottomrule
\end{tabular}
\end{table}

\textbf{Different Modules.}
We also conduct ablation experiments with some components disabled or removed, and the results are displayed in Table~\ref{tblAbl}. We test three different combinations of the channel attention module and the use of the predicted mask from the previous frame. If we remove our channel attention module, the IoU score and the contour accuracy are 3.4$\%$ and 3.7$\%$ lower than the full combination, respectively. Therefore, we can conclude that the channel attention module can strengthen the feature representation to help our network better adapt to foreground pixels. On the other hand, if we take out the predicted mask from the previous frame, the IoU score and the contour accuracy are 5.3$\%$ and 4.8$\%$ lower than the full combination, respectively, which proves that the predicted mask from the previous frame can guide our network to segment the foreground object. Overall, the full NPMCA-net achieves the best performance. It demonstrates that the channel attention module and the use of the predicted mask for the previous frame benefit from each other. 

\begin{table}[width=.9\linewidth,cols=5,pos=h]
\caption{Network module analysis on DAVIS-2017 validation set. `CM' denotes to the channel attention module, and `PM' denotes that the input of current frame with the predicted mask from the previous frame.}\label{tblAbl}
\begin{tabular}{c|cc|cc}
\toprule
   & CM & PM &$\mathcal{J} ($\%$)$ & $\mathcal{F} ($\%$)$ \\
 \hline
1 &  & \checkmark & 68.8 & 73.7 \\
2 & \checkmark &  & 66.9 & 72.6 \\
\hline
3& \checkmark & \checkmark & \textbf{72.2} & \textbf{77.4} \\
\bottomrule
\end{tabular}
\end{table}

\textbf{Encoder Setting.}
Finally, we conduct the ablation study on the setting of encoders with only training with DAVIS-2017 dataset. we conduct the experiment to show the necessity of the parameter-shared encoder for the two references and different encoder for the target frame. The results is shown in Table~\ref{tblenc}. `One encoder' denotes to use same encoder for the three inputs and `Two encoders' denotes to parameter-shared setting. It can be found that with only one encoder, the result is almost 5$\%$ lower than the two-encoder setting. VOS aims to segment the target object from the first frame to the end. To capture consistent reference object feature information, we set parameter-shared encoder for the first frame and previous frame (where background is masked out). Parameter-shared can map the input reference features into the same representation space, thereby the two reference frames’ information can be equally treated. Additionally, parameter-shared can reduce parameters for training. If we use just one encoder for the first, the previous and the target frames, the network will be confused, because the encoder for the current frame needs to encode both image and previous predicted mask information, where the background is not masked out. However, for the first and the previous frames, the background is masked out, and we only use the foreground pixels of the frames.

\begin{table}[width=.9\linewidth,cols=3,pos=h]
\caption{Encoder settings analysis on DAVIS-2017 validation set. `One encoder' denotes to using same encoder for all the inputs `Two encoders' denotes to the setting of parameter-shared only for the reference frames.}\label{tblenc}
\begin{tabular}{c|cc}
\toprule
Encoder Setting & $\mathcal{J} ($\%$)$ & $\mathcal{F} ($\%$)$ \\
\hline
One encoder & 34.7 & 38.6 \\
Two encoders & 41.0 & 43.9 \\
\bottomrule
\end{tabular}
\end{table}

\subsection{Limitations}
\begin{figure}
    \centering
    \includegraphics[width=3.0in]{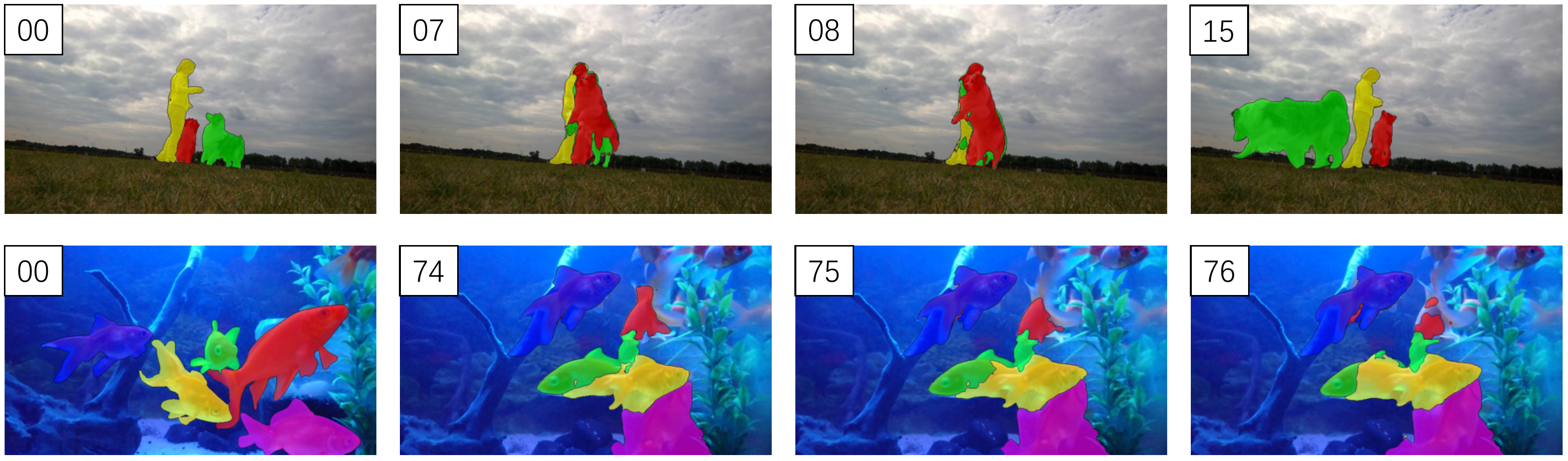}
    \caption{Limited Cases of Our Network}
    \label{fig:failure}
\end{figure}
Some failure cases from our model are shown in Fig.~\ref{fig:failure}. When foreground objects are overlapped, our model tends to produce incorrect segmentation for those occluded objects, especially when the overlapped objects are with the same category. Nevertheless, if the foreground objects are well separated afterwards, our model can adjust to the correct tracking and segmentation status due to the use of the first frame information, like in row 1 of Fig.7. This example shows that our method can catch back to the target object after occlusion. However, when there is occlusion for multi-objects, especially when the targets are in the same category, our method will be confused and lose the target (like in the second row of Fig. 7). To overcome this limitation, we consider that we can generate some prototypes to represent each object and push away their feature distances to make the network be sensitive to different object in the future.

\section{Conclusion}
In this work, we have proposed a new video object segmentation network NPMCA-net, which combines a non-local pixel-matching module and a channel attention module in series connection. Our network achieves the state-of-the-art performance on both DAVIS-2017 and DAVIS-2016 validation set. Additionally, our NPMCA-net has a good generalization ability. Moreover, our network does not need any post-processing, so as to keep a good balance between accuracy and  efficiency. In the future,we consider that we can generate some prototypes to represent each object and push away their feature distances to make the network be sensitive to different object.

\bibliographystyle{elsarticle-num}

\bibliography{reference}


\end{document}